\documentclass[11pt,a4paper]{article}

\usepackage{a4wide}
\usepackage[utf8]{inputenc}
\usepackage[english]{babel}
\usepackage{authblk}
\usepackage{algorithm}
\usepackage{algorithmic}
\usepackage{hyperref}
\usepackage{tikz}
\usepackage{epigraph}
\usepackage[shortlabels]{enumitem}
\usepackage{xcolor,colortbl}
\usepackage{array, makecell, multirow, tabularx}
\usepackage{multicol}
\usepackage{threeparttable}
\usepackage{rotating}
\usepackage{nicematrix}
\usepackage{adjustbox}
\usepackage{framed}
\usepackage{color,soul}
\usepackage[most]{tcolorbox}
\usepackage{fontawesome}
\usepackage[group-separator={,}]{siunitx}
\usepackage{amssymb} 
\usepackage{booktabs}
\setul{0.5ex}{0.3ex}

\usepackage[maxnames=6]{biblatex}
\newtcbox{\badge}[1][red]{
  on line, 
  arc=4pt,
  colback=#1!20!purple,
  colframe=#1!20!purple,
  fontupper=\color{white},
  boxrule=1pt, 
  boxsep=0pt,
  left=3pt,
  right=3pt,
  top=1pt,
  bottom=1pt
}

\definecolor{purple}{HTML}{F542E9}
\usepackage{etoolbox}

\makeatletter
\patchcmd{\epigraph}{\@epitext{#1}}{\itshape\@epitext{#1}}{}{}
\makeatother

\newcolumntype{Y}{>{\centering\arraybackslash}X}

\newcolumntype{C}{>{\raggedright\arraybackslash}p{6.5cm}}
\newcolumntype{A}{>{\raggedright\arraybackslash}p{5cm}}

\newcolumntype{H}{>{\raggedright\arraybackslash\hspace{0pt}}p{3cm}}

\newcolumntype{T}{>{\raggedright\arraybackslash}X}

\AtBeginDocument{%
  \providecommand\BibTeX{{%
    \normalfont B\kern-0.5em{\scshape i\kern-0.25em b}\kern-0.8em\TeX}}}

\begin{document}

\title{Large Language Models for the Automated Analysis of Optimization Algorithms}


\author[1]{Camilo Chacón Sartori}
\author[1]{Christian Blum}
\author[2]{Gabriela Ochoa}
\affil[1]{Artificial Intelligence Research Institute (IIIA-CSIC), Bellaterra, Spain}
\affil[2]{University of Stirling, Stirling, UK}
\affil[ ]{\textit{\{cchacon,christian.blum\}@iiia.csic.es}, \textit{gabriela.ochoa@stir.ac.uk}}


\maketitle

\begin{abstract}
The ability of Large Language Models (LLMs) to generate high-quality text and code has fuelled their rise in popularity. In this paper, we aim to demonstrate the potential of LLMs within the realm of optimization algorithms by integrating them into STNWeb. This is a web-based tool for the generation of Search Trajectory Networks (STNs), which are visualizations of optimization algorithm behavior. Although visualizations produced by STNWeb can be very informative for algorithm designers, they often require a certain level of prior knowledge to be interpreted. In an attempt to bridge this knowledge gap, we have incorporated LLMs, specifically GPT-4, into STNWeb to produce extensive written reports, complemented by automatically generated plots, thereby enhancing the user experience and reducing the barriers to the adoption of this tool by the research community. Moreover, our approach can be expanded to other tools from the optimization community, showcasing the versatility and potential of LLMs in this field.
\end{abstract}

\begin{figure}[!h]
  \includegraphics[width=\linewidth]{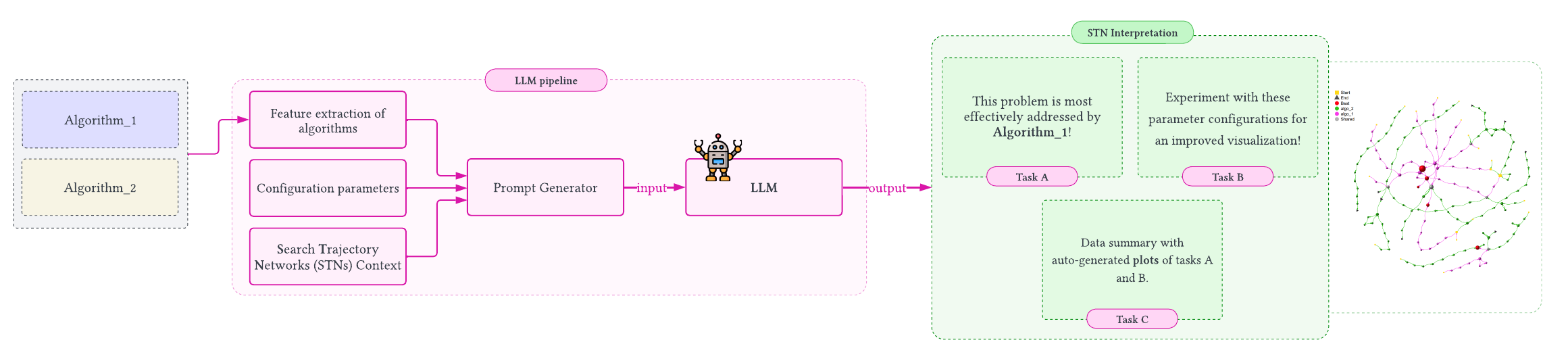}
  \caption{Graphical overview on the automation of the analysis of optimization algorithm behavior using Large Language Models (LLMs).}
  \label{fig:teaser}
\end{figure}

\section{Introduction}

\epigraph{GPT-4 can serve as an advanced natural language processing layer within STNWeb, offering intuitive explanations and insights based on complex data visualizations to enhance user understanding and interaction with the tool.}{--- \textup{GPT-4-turbo}\footnote{Generated from the prompt: ``Please provide us with a one-sentence definition of GPT-4's role in integration to improve explainability in our visualization tool (STNWeb)."}
}

For a tool to be effective and widely adopted, it must be user-friendly. A research tool is designed for experts, and the traditional method of learning how to use it requires guidance and explanations from an expert. This often involves an extensive literature study, which can be a time-consuming task. Recently, in the field of optimization, the use of visualization tools has been proposed to enhance the analysis of algorithms such as metaheuristics~\cite{gendreau2005metaheuristics, blum2003metaheuristics}. An example are Search Trajectory Networks (STNs), which enable the exploration of the behavior of various optimization algorithms on a specific problem, be it discrete/combinatorial or continuous~\cite{ochoa2020search, OCHOA2021107492}. The value of this approach lies in its ability to reveal how a given problem \textit{influences} the behavior of an algorithm, thereby enabling researchers to deepen their understanding and make more informed design choices when developing new algorithms. The introduction of the web version of STNs, STNWeb~\cite{CHACONSARTORI2023100558}, constituted a big step towards the automation 
of the process of generating STNs. However, this simplified visualization process must be accompanied by guidance for the interpretation of the obtained graphics. For this purpose, in this paper, we examine the use of Large Language Models (LLMs).

LLMs have recently transformed the field of natural language processing (NLP) in a significant way by delivering high-quality responses, making them indispensable for professionals across various disciplines~\cite{Qiu_2020}. Models like OpenAI's ChatGPT and DALL-E~2 have set new benchmarks in industry~\cite{openai2023gpt4, ramesh2021zeroshot}. Furthermore, GitHub Copilot has demonstrated the capability of LLMs to automatically generate code~\cite{Sun2022-eh, 10.1145/3491101.3519665}, significantly streamlining the software development process. Beyond their practical applications, these models have also started to show a capacity for designing new algorithms, particularly in the domain of metaheuristics~\cite{10.1145/3583133.3596401}. Additionally, LLMs have shown promise in various other applications, such as sentiment analysis~\cite{zhang2023sentiment}, text classification~\cite{sun2023text}, machine translation~\cite{zhang2023prompting}, and question answering systems~\cite{kamalloo2023evaluating}. However, LLMs also come with their challenges. One major issue is that their responses can sometimes include so-called hallucinations, or unrealistic data, which can be problematic in real-world applications. To address this, techniques such as prompt engineering have emerged as a solution~\cite{Reynolds2021-eh}. By carefully designing the input prompts, these techniques can help reduce the likelihood of hallucinations in LLMs, making their responses more reliable and trustworthy~\cite{feldman2023trapping}. Another significant advantage of prompt engineering is that it enhances the explainability of LLMs~\cite{zhao2023explainability}. By breaking down complex queries into simpler and more focused prompts, it becomes easier to understand how the model arrived at its response. Thus, prompting can be viewed as \textit{natural language programming}.

In this paper, we make use of LLMs within STNWeb to automatically generate prompts for the user. This functionality will help users to better understand the generated graphics and their results, while also increasing their familiarity with the algorithms' behavior when applied to specific optimization problems. Additionally, we will demonstrate that LLMs are capable of producing code that generates basic plots (e.g., bar plots), further enhancing the natural language report provided by the LLM. An overview of the integration process consists of three distinct tasks: A, B, and C, as shown in Figure~\ref{fig:teaser}. The paper unfolds as follows. Section~2 introduces the fundamental concepts of STNs and the role of LLMs in enhancing explainability. The discussion proceeds in Section~3 with an exploration of the incorporation of LLMs into STNWeb, including a detailed description of how prompt templates are constructed. Section~4 delves into an empirical assessment, presenting the evaluation process for the efficacy of these templates for various prompts, as well as the experimental setup and the methodology employed for analyzing the outcomes. Section~5 discusses the potential applications of LLMs within the domain of optimization tools. The paper concludes by summarizing our findings and suggesting future directions for augmenting the explainability and trustworthiness of STNWeb concerning the integration of LLMs.

\section{Background}

\subsection{Search Trajectory Networks (STNs)}

STNs are visualizations of directed graphs designed to analyze stochastic optimization algorithms such as metaheuristics. Their purpose is to equip researchers with the means to gain a deeper understanding of the behavior exhibited by multiple algorithms when applied to a specific instance of an optimization problem. This visualization helps discern, for example, whether an algorithm is attracted by a local optimum or exhibits limited exploration capabilities. Consequently, the explainability of STNs relies on our understanding of such a visualization. The next paragraph will describe how to make an STN graph easier to understand by describing the concept of explainability in this context.

\begin{figure}[!h]
\centering
\begin{center}
  \includegraphics[width=0.6\linewidth]{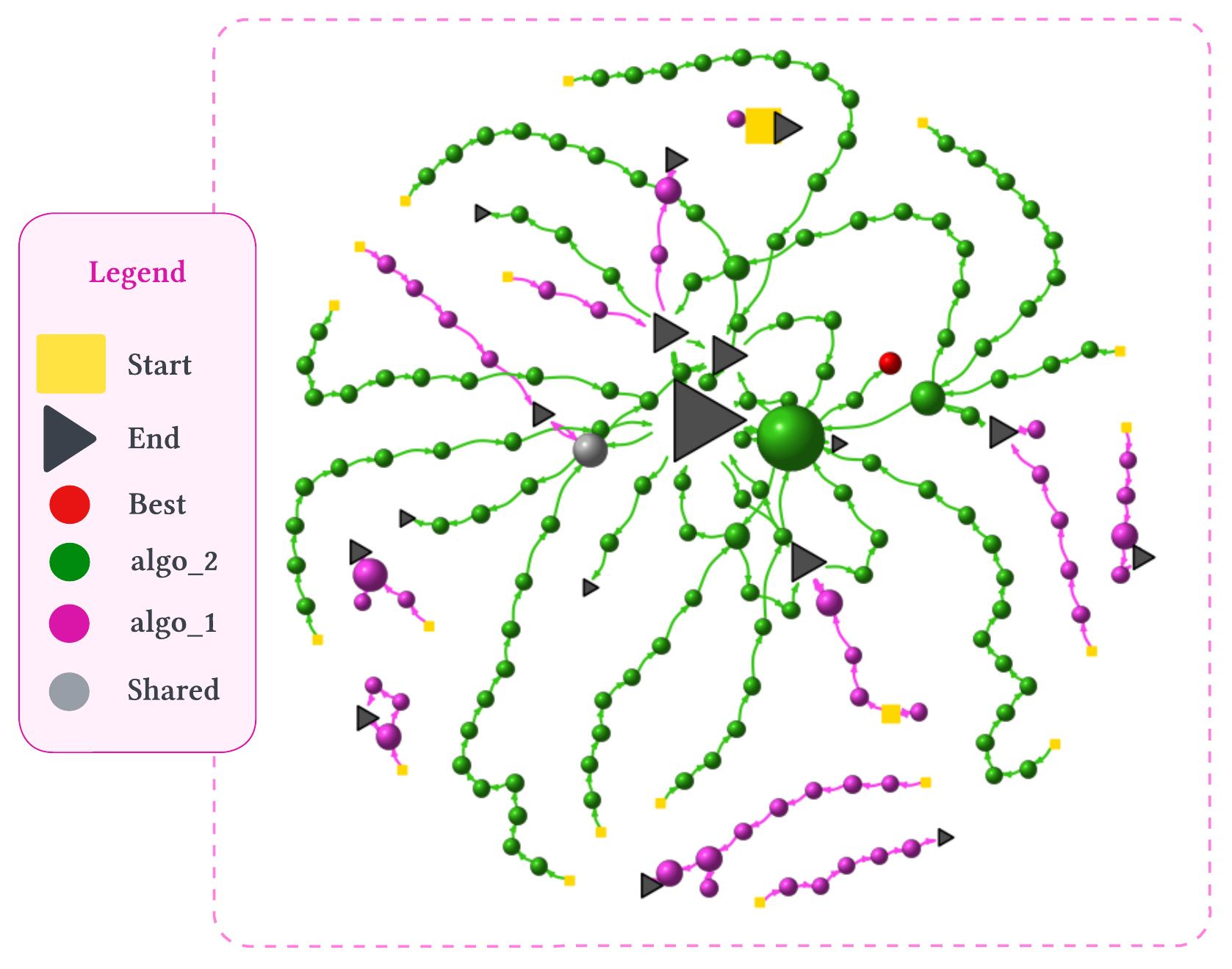}
  \caption{Example of an STN produced by STNWeb.}
  \label{fig:stn-example}
\end{center}
\end{figure}
Figure~\ref{fig:stn-example} provides an example of an STN graphic, demonstrating the performance of two different optimization algorithms for the \emph{Rastrigin function}, a challenging continuous optimization problem. The illustration displays the trajectories resulting from 10 runs of each algorithm. Each dot (vertex) of the STN represents, in general terms, a chunk of the search space containing at least one solution. These chunks of the search space are produced by search space partitioning schemes such as agglomerative clustering. The graphical elements in such an STN have the following meaning:
\begin{itemize}
    \item Different algorithms' trajectories are depicted in distinct colors, detailed in the legend. For example, in Figure~\ref{fig:stn-example}, the 10 trajectories of \textsc{algo\_1} are shown in \textcolor{purple}{purple}, while \textsc{algo\_2}'s trajectories are in \textcolor{green}{green}.\footnote{In this context, understanding the nature of these algorithms is not crucial.}
    \item Initial points are marked with yellow squares. 
    \item Trajectory endpoints are represented as dark grey triangles, respectively by red dots. Dark grey triangles indicate endpoints not corresponding to the best-found solutions, while red dots indicate endpoints corresponding to the best-found solutions. Note that one trajectory of \textsc{algo\_2} in Figure~\ref{fig:stn-example} ends in a best-found solution.
    \item Pale grey dots represent solutions (resp.~chunks of the search space) shared across trajectories of at least two different algorithms.
    \item Vertex or dot size indicates the count of algorithm trajectories passing through it: larger vertices indicate more traversing algorithm trajectories.
\end{itemize}

In particular, the STN graphic of Figure~\ref{fig:stn-example} reveals the presence of an attraction area within the search space, particularly for \textsc{algo\_2}. Seven out of 10 \textsc{algo\_2} trajectories are notably drawn toward the region in the middle of the graphic. In contrast, \textsc{algo\_1} is not attracted by that area. Moreover, no overlap among \textsc{algo\_1}'s trajectories can be detected, which indicates a low robustness. 

Note that this simple example already provides valuable insights into algorithm performance. However, while the visualization expedites the analysis of algorithm performance in contrast to traditional result table examinations, comprehending the “language” of the STN context requires prior knowledge. Researchers likely have to go through the initial article introducing STNs~\cite{OCHOA2021107492} and its subsequent ones~\cite{10.1145/3587828.3587843, CHACONSARTORI2023100558}, a process that can be notably time-consuming. The aspiration to eliminate the prerequisite of prior knowledge for interpreting STN graphics aims to enhance and simplify the tool. Therefore, in this paper, we integrate LLMs to automatically produce a natural language report and basic graphics (as supplemental aids) to facilitate the interpretation of STN visualizations.

\begin{table}[!t]
\centering
\caption{Guidelines and expected inferences for each task, specifying the LLMs' roles and responsibilities.}
\label{table:tasks}
\setlength\extrarowheight{9.6pt}
\settowidth\rotheadsize{\textsc{Task A}}

\scalebox{0.63}{
\begin{tabularx}{\textwidth}{l H lC A}

\rowcolor[HTML]{ffd6f5}
\thead[b]{Prompt\\Template} 
    &   \thead[b]{Tasks}
        & \multicolumn{2}{c}{Rule/Instruction within the prompt}
                & \thead[b]{Inference\\Expected}         \\
\multirow{12}{*}{\rotcell{\textsc{Task A}}}
    &  \multirow{12}{=}{Determine if there is a clear winner among the compared algorithms}
        &   \multirow{4}{*}{1}
            &   \multirow{3}{=}{\\ \textit{The more nodes point to nodes of the best fitness (this does not assume that it represents the global optimum), the higher the algorithm's quality because it can obtain a better result.}}
                &   \multirow{12}{=}{With three rules and the values automatically computed by STNWeb from the input files of each algorithm, the LLM is requested to come up with a conclusion. It is supposed to identify the best algorithm by providing its name. Moreover, if multiple algorithms exhibit similar values for all features, the LLM should acknowledge their similarity and indicate the inability to determine a clear winner.}        \\ 
    &     &   &   &                                               \\
    &     &   &   &                                               \\
    &     &   &   &                                               \\
    &     & \multirow{3}{*}{2} &   \multirow{3}{=}{\textit{The algorithm that has more inter-trajectory connectivity is likely to be more robust. If and only if it finds nodes of the best fitness.}}   \\
    &     &   &   &                                               \\
    &     &   &   &                                               \\
    &     &\multirow{4}{*}{3}  &   \multirow{3}{=}{\textit{For a minimization problem, indicating that an algorithm is superior involves favoring a smaller average fitness value. Whereas in the case of maximization, declaring an algorithm as better necessitates a higher average fitness value.}}     \\
    &     &   &   &                                               \\
    &     &   &   &                                               \\
    &     &   &   &                                               \\
    &     &   &   &                                               \\
    \cline{1-5}                                        
\multirow{19}{*}{\rotcell{\textsc{Task B}}}
    &  \multirow{19}{=}{Suggest a better parameter setup for the agglomerative clustering algorithm to produce enhanced STN graphs}
        &   \multirow{3}{*}{1}
            &   \multirow{3}{=}{\textit{Cluster size (percentage): Maximal cluster size in terms of the percentage of all solutions a cluster contains.}}
                &   \multirow{18}{=}{Our expectation in this scenario is that the LLM, equipped with information about four parameters controlling the search space partitioning feature of SNTWeb, together with the user-selected parameter values, is able to propose better parameter values based on the context and the provided instructions.}        \\
    &     &   &   &                                               \\
    &     &   &   &                                               \\
    &     & \multirow{3}{*}{2} &   \multirow{3}{=}{\textit{Volume size (percentage): Maximal cluster size in terms of the percentage of the covered search space volume spanned by the solutions a cluster contains.}}   \\
    &     &   &   &                                               \\
    &     &   &   &                                               \\
    &     &   &   &                                               \\
    &     &\multirow{3}{*}{3}  &   \multirow{3}{=}{\textit{Distance measure: A function that measures the distance between solutions, influencing the creation of clusters. Possible values: Hamming, Euclidean, Manhattan.}}     \\
    &     &   &   &                                               \\
    &     &   &   &                                               \\   
    &     &   &   &                                               \\
     &     &\multirow{6}{*}{4}  &   \multirow{3}{=}{\textit{Cluster number: Number of clusters obtained for these solutions (from lowest to highest partitioning). The maximum number implies no partitioning, while lower values result in increased partitioning. Good results are obtained when the cluster number is above the minimum value but far from the maximum.}}     \\
    &     &   &   &                                               \\
    &     &   &   &                                               \\
    &     &   &   &                                               \\
    &     &   &   &                                               \\
    &     &   &   &                                               \\
    &     &   &   &                                               \\
    \cline{1-5}     
\multirow{8}{*}{\rotcell{\textsc{Task C}}}
    &  \multirow{8}{=}{Produce an automatically generated summary with plots for tasks A and B}
        &   \multirow{3}{*}{1}
            &   \multirow{3}{=}{\textit{Generate a grouped bar plot, considering both the best-performance (in sky blue) and the average-performance (in orange).}}
                &   \multirow{8}{=}{The second instruction requests the suggestion of parameters for \textsc{Task B} to showcase parameter changes. This highlights the LLM's ability to automatically create graphs through inference, utilizing prompts and a tabular data file generated by STNWeb to specify columns for plot generation.}
                \\
    &     &   &   &                                               \\
    &     &   &   &                                               \\
    &     & \multirow{4}{*}{2} &  \multirow{3}{=}{\textit{Generate a grouped bar plot with the X-axis representing the old\_configuration and new\_configuration, and the Y-axis representing cluster-size (in sky blue), volume-size (in orange), and cluster-number (in purple).}}   \\
    &     &   &   &                                               \\
    &     &   &   &                                               \\
    &     &   &   &                                               \\
    &     &   &   &                                               \\
    \cline{1-5}
\end{tabularx}
}
\end{table}

\subsection{Large Language Models (LLMs)}

LLMs, such as GPT-4~\cite{openai2023gpt4}, Mixtral~\cite{jiang2023mistral}, Gemini~\cite{geminiteam2023gemini}, and Llama~2~\cite{touvron2023llama}, are sophisticated language models involving a huge number of parameters and exhibiting a remarkable ability to learn. They enable the comprehension and generation of human language. Many tools---such as ChatGPT, Bing Chat, and Google Bard---have proven the practical value of these models.


At the heart of LLMs lies the self-attention mechanism within transformers~\cite{vaswani2023attention}, which acts as the fundamental unit for making reliable predictions for each token in language sequences. Transformers have revolutionized the realm of natural language processing (NLP) with their proficiency in effectively completing text sequences. This innovation leverages the power of modern high-performance hardware parallelization while also grasping an extensive array of textual dependencies. This dual capability enhances the contextual quality of the generated text. As the number of parameters scales massively in emerging LLMs, various abilities have come to light~\cite{wei2022emergent}, including few-shot in-context learning~\cite{brown2020language}, zero-shot problem-solving~\cite{kojima2023large, sanh2022multitask}, chain of thought reasoning~\cite{wei2023chainofthought}, instruction following~\cite{ouyang2022training}, directional stimulus~\cite{li2023guiding}, and retrieval-augmented generation~\cite{lewis2021retrievalaugmented}. These abilities can be utilized through the prompt, serving as the entry point for engaging with any LLM. The LLM may produce varying responses based on the specific problem at hand. In fact, selecting the appropriate ability is essential, and this process is referred to as prompt engineering~\cite{Zhou2022-tx}.

In our approach, we will employ a pre-trained model because it enables us to attain our objectives by simply adjusting the prompts~\cite{Qiu_2020, liu2021pretrain}. This approach is advantageous as it bypasses the need to modify the weights of the training set, which is necessary in the fine-tuning process. Additionally, we eliminate the requirement of having our own LLM, which requires substantial computing power. Instead, we will utilize off-the-shelf LLMs, including paid options such as OpenAI's GPT-4, as well as freely available LLMs that operate on platforms like Hugging Face.

Therefore, by selecting suitable emergent abilities from the LLM, we will develop three distinct prompts. Each one is designed to carry out a specific function aimed at providing information to the STNWeb user. This approach will facilitate STNWeb's use and will significantly reduce the need for the user to have pre-existing knowledge of the interpretation of STN graphics.

\section{Integrating LLMs into STNWeb}


Our goal is to utilize LLMs to generate interpretations of STN graphics for supporting users without specialized knowledge. For this purpose, STNWeb was extended to interact with LLMs by generating prompts, and by displaying the interpretations received from the LLMs (that is, natural language summaries and additional plots) for the users' benefit. This streamlines STNWeb into a more user-friendly and effective analysis tool, accessible to both beginners and experts.

The integration of STNs and LLMs consists of two stages. In the first phase, the prompt templates are engineered. In the next phase, some basic features of the STN graph are extracted for each algorithm. The parameters used for search space partitioning by agglomerative clustering are also extracted. This step aims to gather data on the behavior of each algorithm and the partitioning method.  

STNWeb autonomously executes both steps without the need for user engagement. It extracts features from the provided input files (containing raw data about the algorithm trajectories) and creates prompts according to the results produced by each algorithm. Subsequently, the LLM produces a tailored interpretation. In the following, we outline the specifics of these two stages.

\subsection{Prompt Engineering}

Prompting is an intuitive and user-friendly interface for humans to interact with generalist models such as LLMs. However, guiding LLMs effectively requires careful engineering. Especially, these models struggle to interpret prompting in a manner equivalent to human understanding~\cite{DBLP:journals/corr/abs-2109-01247}. This has led some researchers, like~\cite{Reynolds2021-eh}, to assert that prompt engineering can be regarded as a form of \textit{programming in natural language}.

To incorporate LLMs into STNWeb, we created three prompt templates referred to as Tasks A, B, and C, which will leverage the information gathered from the STN graphic to be analyzed, to handle the following tasks:

\begin{enumerate}[(A)]
    \item Determine if there is a winner among the compared algorithms!
    \item Suggest (if necessary) a better parameter setup for the agglomerative clustering algorithm, to produce STN graphics that are easier to interpret!
    \item Produce an automatically generated summary with plots for tasks A and B!
\end{enumerate} 

\begin{figure*}[t]
\centering
\begin{center}
  \includegraphics[width=\linewidth]{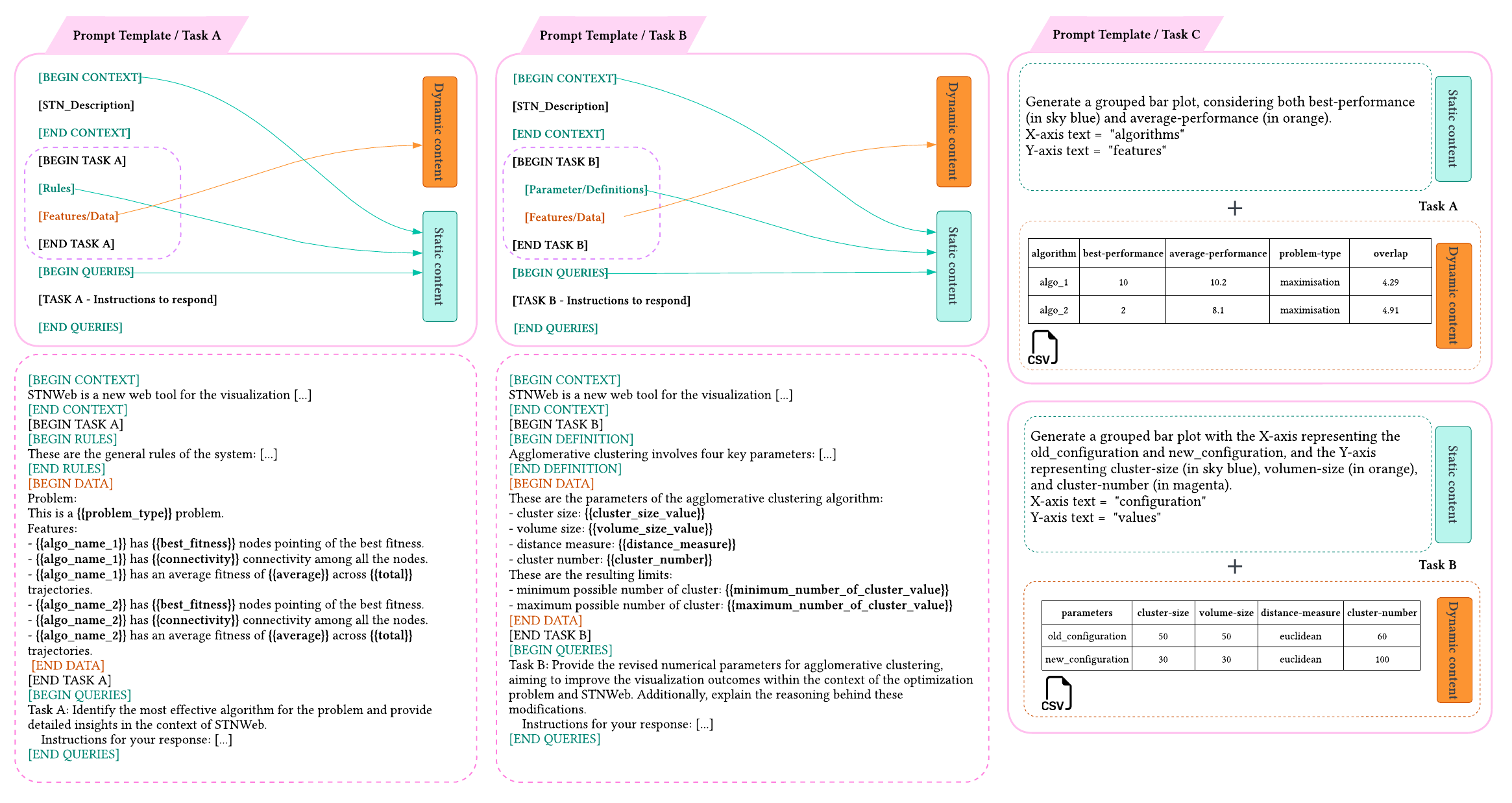}
  \caption{Prompt templates that are automatically generated in STNWeb for each task.}
  \label{fig:prompt-templates}
\end{center}
\end{figure*}

These templates are automatically instantiated by STNWeb and do not require user intervention. As shown in Figure~\ref{fig:prompt-templates}, templates for Tasks A and B feature tags that include a block of text linked to distinct instructions for the respective task. These instructions can be categorized as static, remaining unchanged in each new prompt (as indicated in cyan color), or dynamic, altering based on STN features and user parameters (orange tags). Implementing such a structured format aims to prevent LLMs from succumbing to hallucinations and to enhance the coherence of the responses. We will now proceed to clarify the syntactic structure of the templates associated with tasks A, B, and C.

\subsubsection{Task A} In the first column of Figure~\ref{fig:prompt-templates}, the task~A prompt template begins with the \textsc{[CONTEXT]} tag, providing a concise description of STNWeb. This causes the LLM to respond considering the terminology employed in the description. Next, the \textsc{[RULES]} tag outlines the criteria established for evaluating the superiority of one algorithm over another. The current set of rules is provided in the 3rd column of Table~\ref{table:tasks}. The next tag, \textsc{[DATA]}, is different from the previous ones because it is dynamic, containing values of three algorithm-specific features extracted from the STN.  The template ends with the inclusion of the \textsc{[QUERIES]} tag, which provides the LLM with guidelines on crafting responses, encompassing aspects such as the formatting of word structures. This information will help in thoroughly evaluating the responses of each tested LLM.

\subsubsection{Task B} In the central column of Figure~\ref{fig:prompt-templates}, analogous to task~A, the template starts with the \textsc{[CONTEXT]} tag, providing the context of STNWeb. Afterward, the \textsc{[PARAMETERS DEFINITIONS]} tag outlines the parameters of the search space partitioning method (agglomerative clustering). In particular, this includes four parameters as outlined in the third column of Table~\ref{table:tasks}. Next, the template \textsc{[DATA]} tag encapsulates the user-selected configuration of these parameters for producing the STN graphic to be interpreted. The template ends with the \textsc{[QUERIES]} tag, with the same aim as in task~A.

\subsubsection{Task C} The right column of Figure~\ref{fig:prompt-templates} shows the prompt devised for this task. Distinct from the preceding tasks, Task~C encompasses two concise templates, each one aims to prompt the LLM to generate a plot following the given instructions (the next section elaborates on our approach to achieving this). To enable the LLM to discern the data for each plot generation, a tabular data file (in CSV format) is appended to each prompt. The first template instructs the LLM to generate a grouped bar plot elucidating the characteristics of each algorithm, referencing the column names in the file (linked to task~A). The second template directs the LLM to generate a grouped bar plot delineating the parameter settings chosen in agglomerative clustering for generating the STN graphic, and the new settings recommended by the LLM. Consequently, this prompt relies on the response from task~B, with the column names explicitly indicated in the prompt. \\

See Table~\ref{table:tasks} for the STNWeb-specific rules/instructions that are associated with each task from a semantic perspective, as well as the anticipated inferences we expect from the LLM. Keep in mind that the instructions provided in the templates are currently quite basic. However, as we will show in the results section, they have been instrumental in serving as a proof of concept for the use of LLMs for STN graphic interpretation. 

It is also important to keep in mind that these prompts are designed for zero-shot learning~\cite{kojima2023large, sanh2022multitask}, meaning that no decision examples are provided for any of the tasks. In simpler terms, we do not offer input-output examples within the prompt to guide the model.
We anticipate that the model will be capable of making decisions based solely on the rules outlined in the description, without any explicit examples. This contributes to a more streamlined and manageable prompt design, only requiring a clearly defined set of rules and parameters associated with STNWeb.

\subsection{Feature Extraction}

The dynamic aspects of the devised prompt templates can be found within the \textsc{[DATA]} tag for tasks A and B, and in the CSV files that are attached to task~C. Next, we describe the process of extracting features for tasks A, B, and C. Specifically, we explain how features are derived from the data (search trajectories) of the algorithms (task~A), the selected configuration of STNWeb (task B), and the input file format (task C).

\subsubsection{Task A} STNWeb currently extracts three essential features that are typically used by STNWeb users to identify the better-performing algorithm. Our objective here is to empower the LLM with the ability to automatically determine the winning algorithm by providing it with these features. If the LLM cannot confidently determine a winner, it will return a [draw] to indicate that no clear preference can be established for any of the algorithms in the comparison.

Next, we provide the definitions of the three features that are extracted by STNWeb to build the prompt. Moreover, the sentence to be included in the prompt is indicated. Hereby, the dynamic numerical values are highlighted in \textcolor{green}{green}, while the keywords underlined in \textcolor{purple}{purple} aid the LLM in understanding the pre-established rules. The algorithm's name is underlined in black. \\

\textbf{Feature 1: total best global fitness}. This feature counts the number of trajectories of each algorithm \textsc{algo\_k} that end in a solution whose fitness (objective function value) is equal to that of the best solution found by all algorithm trajectories in the comparison. We consider that the higher this number, the higher the chance that this algorithm is the winner of the comparison.
\begin{equation}
\resizebox{.9\hsize}{!}{
$
    \begin{split}
        \text{TotalBestGlobalFitness}_{\text{\textsc{algo\_k}}} = \sum_{i=1}^{|M_k|} \mathbb{I}(\text{BestFitness}_{k,i} = \text{BestGlobalFitness}),
    \end{split}
$
}
\label{eq:globalbestfitness}
\end{equation}
where $M_k$ is the set of trajectories of algorithm \textsc{algo\_k}, the term $\text{BestGlobalFitness}$ stores the highest fitness value found in all algorithm trajectories, $\text{BestFitness}_{k,i}$ refers to the fitness of the best solution of the $i$-th trajectory of \textsc{algo\_k}, and function $\mathbb{I}(\cdot)$ is the indicator function that outputs 1 if the condition within the parenthesis is true, and 0 otherwise.

For example, the value of this feature for algorithm \textsc{algo\_2} from the example in Figure~\ref{fig:stn-example} is 1. Therefore, the following sentence will be added to the prompt.

\begin{tcolorbox}[colback=purple!5, colframe=gray!10!gray,  boxrule=0.5pt]

\textit{\underline{\textsc{algo\_2}} has \setulcolor{green}\ul{1} nodes pointing to nodes with the \setulcolor{purple}\ul{best fitness}.}
\end{tcolorbox}
\textbf{Feature 2: inter-trajectory connectivity}. Connectivity arises between trajectories of the same algorithm when solutions from distinct trajectories fall into the same node of the STN to be analyzed. In the current rule system, we hypothesize that an algorithm that shows more interconnected components between its algorithm trajectories is likely to be a rather robust algorithm, even though this is not always true.


\begin{equation}
\text{Connectivity}_{\text{\textsc{algo\_k}}} = \frac{\text{\# pairs } T \not= T' \in M_k \text{ with an overlap}}{(|M_k| \cdot (|M_k| - 1))/2},
\label{eq:connectivity}
\end{equation}
where $M_K$ is the set of trajectories of algorithm \textsc{algo\_k}, and $T$ and $T'$ are trajectories from $M_k$. Thus, provided that Equation~\ref{eq:connectivity} yields a value of 0.62, the sentence to be generated for the prompt is as follows.

\begin{tcolorbox}[colback=purple!5, colframe=gray!10!gray, boxrule=0.5pt]
\textit{\underline{\textsc{algo\_2}} has \setulcolor{green}\ul{0.62} \setulcolor{purple}\ul{connectivity} among all the nodes.}
\end{tcolorbox}

\textbf{Feature 3: average fitness}. The term average fitness refers to the average of the objective function values of the last solutions of all trajectories of an algorithm. We include this feature because, generally, if an algorithm \textsc{algo\_1} has an average fitness that is better than the one of an algorithm \textsc{algo\_2}, \textsc{algo\_1} is likely to be the best algorithm.

\begin{equation}
\text{AvgFitness}_{\text{\textsc{algo\_k}}} = \frac{1}{|M_k|} \sum_{i=1}^{M}\text{BestFitness}_{k,i}
\label{eq:avgallbestfitness}
\end{equation}

This equation computes the average fitness of algorithm \textsc{algo\_k} over its $|M_k|$ trajectories. The term $\text{BestFitness}_{k,i}$ refers to the best fitness value obtained by algorithm \textsc{algo\_k} in the $i$-th trajectory. 

Finally, in this scenario, in case Equation~\ref{eq:avgallbestfitness} returns 78.081 and there are 10 trajectories, the sentence to be produced for the prompt is as follows.

\begin{tcolorbox}[colback=purple!5, colframe=gray!10!gray,boxrule=0.5pt]
\textit{\underline{\textsc{algo\_2}} has an \setulcolor{purple}\ul{average fitness} of  \setulcolor{green}\ul{78.081} across \setulcolor{green}\ul{10}  trajectories.}
\end{tcolorbox}

\subsubsection{Task B} Task A gathers features from the input files containing the trajectory data for all algorithms in STNWeb, while task~B retrieves values chosen in STNWeb for agglomerative clustering, which are used for search space partitioning. Due to space constraints, we omit the meaning of each agglomerative clustering parameter in this article. However, Table~\ref{table:tasks} (row referring to task~B), describes the objectives that the rules aim to accomplish. 

This task generates a paragraph as shown below for the corresponding prompt.

\begin{tcolorbox}[ colback=purple!5, colframe=gray!10!gray, boxrule=0.5pt]
\textit{These are the \setulcolor{purple}\ul{parameters of the agglomerative clustering} algorithm:\\
- \setulcolor{purple}\ul{cluster size}: \setulcolor{green}\ul{5\%}\\
- \setulcolor{purple}\ul{volume size}: \setulcolor{green}\ul{5\%}\\
- \setulcolor{purple}\ul{distance measure}: \setulcolor{green}\ul{Euclidean}\\
- \setulcolor{purple}\ul{cluster number}: \setulcolor{green}\ul{400}\\
These are the resulting limits:\\
- \setulcolor{purple}\ul{minimum possible number of clusters}: \setulcolor{green}\ul{207}\\
- \setulcolor{purple}\ul{maximum possible number of clusters}: \setulcolor{green}\ul{{574}
}}
\end{tcolorbox}

Hereby, \textit{cluster size} and \textit{volume size} are limiting factors for the agglomerative clustering algorithm, the \textit{distance measure} and the number of clusters for the STN graphic are also user-selected. Hereby, the value of 400 is chosen by the user from the range $[207, 574]$, resulting from agglomerative clustering.

\subsubsection{Task C} In this task, prompts are static, while the two CSV files are dynamic. The first file holds information related to task~A, and the second file contains details about task~B, including the information returned by the LLM. To provide both a prompt and tabular files to an LLM, we utilize Chat2VIS---a generator of data visualizations that can be used with natural language and is compatible with LLMs such as GPT and Code Llama~\cite{10121440, maddigan2023chat2vis}. This system interprets the prompt and tabular file data, generating a plot through automatically generated Python code. Examples are shown in the right column of Figure~\ref{fig:prompt-templates} and at the bottom of Figure~\ref{fig:examples-tasks}.

LLMs can recognize numbers in context, categorize sentence keywords, and link them to predefined rules (review the words and task rules). However, for complex mathematical reasoning, LLMs may be less effective~\cite{lewkowycz2022solving, wei2023chainofthought}. In our approach, we have opted for straightforward task-specific rules, as this simplicity enhances the LLM's ability to engage in more accurate reasoning when faced with basic inequalities.

\section{Empirical evaluation}

We use an approach that includes both system and human evaluation to evaluate the quality of prompts for tasks A, B, and C. System evaluation will involve comparing actual outputs to expected outputs for tasks A and B, identifying specific word formats indicated in the \textsc{[QUERIES]} tag of each prompt. Human evaluation is provided by STN experts, who assess the clarity, specificity, and effectiveness of the prompts. Human evaluation is essential even when system evaluation is deemed satisfying for all LLMs, as different LLMs may generate text with varying levels of detail, organization, and clarity. We will now discuss the specific setup and methodology used to evaluate each template for every task.

\subsection{Setup}

To conduct the evaluation, we will utilize web platforms that enable the generation of multiple outputs of different LLMs. 

\subsubsection{Tasks~A and~B} To assess tasks~A and~B, we employ Chatbot Arena, an open-source research project developed collaboratively by LMSYS and UC Berkeley SkyLab members~\cite{zheng2023judging}. This platform offers access to over twenty LLMs, including both proprietary and open-source options. For our evaluation, we have selected three LLMs: GPT-4-turbo~\cite{openai2023gpt4}, mixtral-8x7b-instruct-v0.1~\cite{jiang2023mistral}, and tulu-2-dpo-70b~\cite{ivison2023camels}. Chatbot Arena features a Leaderboard that ranks LLMs based on the quality of their responses, as determined by votes from more than \num{200000} users. This useful tool allows us to identify the top-performing LLMs, including GPT-4-turbo, which ranks first on the Leaderboard.\footnote{As of February 2024.} To include also open-source LLMs, we have chosen the next two highest-ranking options with open-source licenses: mixtral-8x7b-instruct-v0.1 and tulu-2-dpo-70b. This selection will provide us with a comprehensive understanding of the performance of various LLMs on tasks~A and~B.

\subsubsection{Task~C} As mentioned above, we use Chat2VIS---a generating data visualizations tool via natural language that offers two distinct LLM categories: those accessible through OpenAI, a wide range of different GPT models~\cite{brown2020language, openai2023gpt4}, and Code Llama~\cite{Roziere2023-yl} developed by Meta. Plot generation is carried out through Python code generation, therefore, Chat2VIS reveals the coding generation ability of its supported LLMs~\cite{10121440, maddigan2023chat2vis}. This capacity is not universal among all LLMs, as noted in a survey by Zheng et al.~\cite{zheng2024survey}.

\begin{figure}[!h]
\begin{center}
  \includegraphics[width=0.8\linewidth]{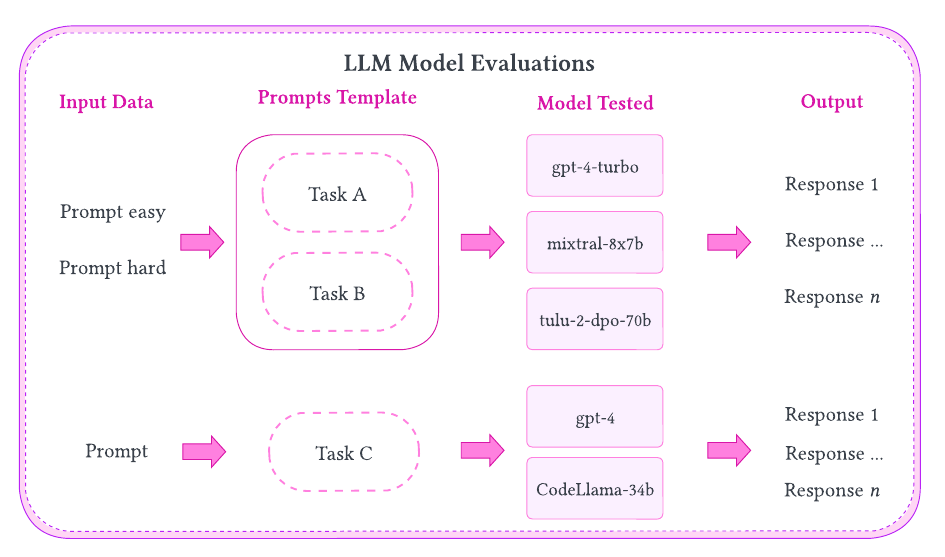}
  \caption{Methodology to evaluate each LLM for every task.}
  \label{fig:evaluation-llm}
\end{center}
\end{figure}

\subsection{Methodology}

We created and pre-labeled four prompts (based on selected STNWeb graphics) with two prompts for each task, A and~B. The difficulty level of these prompts is categorized as either easy or hard, and they will be generated according to the template created for both tasks. The level of difficulty for each prompt is determined by the information found within the dynamic section tag \textsc{[DATA]}. Then, each prompt is provided to each LLM, and the output is evaluated using both system and human evaluation. This evaluation process is executed five times due to the stochasticity of LLMs' output. Task~C will have a single prompt, and only human evaluation since the output (bar plots) is not in the form of natural language text. The methodology described below is also illustrated in Figure~\ref{fig:evaluation-llm}.

\subsubsection{Task A} An easy-type prompt involves comparing the features of two algorithms where one is the clear winner, while a hard-type prompt compares three algorithms where two perform comparably but better than the third algorithm. For instance, when comparing two algorithms named \textsc{algo\_1} and \textsc{algo\_2}, as per the \textsc{[QUERIES]} tag, the LLM should provide the winner in square brackets \textsc{[algorithm\_name]} or indicate a \textsc{[draw]} if their performance is similar. This approach allows us to evaluate whether or not the LLM has made an error. Each prompt will be evaluated using the following scoring function: $LLM_i = \frac{correct}{5}$ for each LLM, where $corrects$ represents the number of LLM outputs deemed to be correct (out of five). In the context of human evaluation, each LLM can only win once; the scoring formula is then $LLM_i = \frac{wins}{5}$ for each model, where $wins$ denotes the total number of victories.

\subsubsection{Task B} Similarly to task A, we will develop two prompts for this task, one simpler and one harder. The evaluation of the system will involve comparing the output of the LLM to the parameter setting provided in the prompt. The LLM output's compliance will be determined based on the rules outlined for this task (see Table~\ref{table:tasks}, second row). Here, as in Task~A, both system and human evaluation will be utilized.

\subsubsection{Task C} As the output of the LLMs in this task is not in the form of natural language text, the validation of the plots' quality (and correctness) will be done only by human evaluation.

\begin{figure}[!th]
\begin{center}
  \includegraphics[width=0.90\linewidth,height=19.2cm]{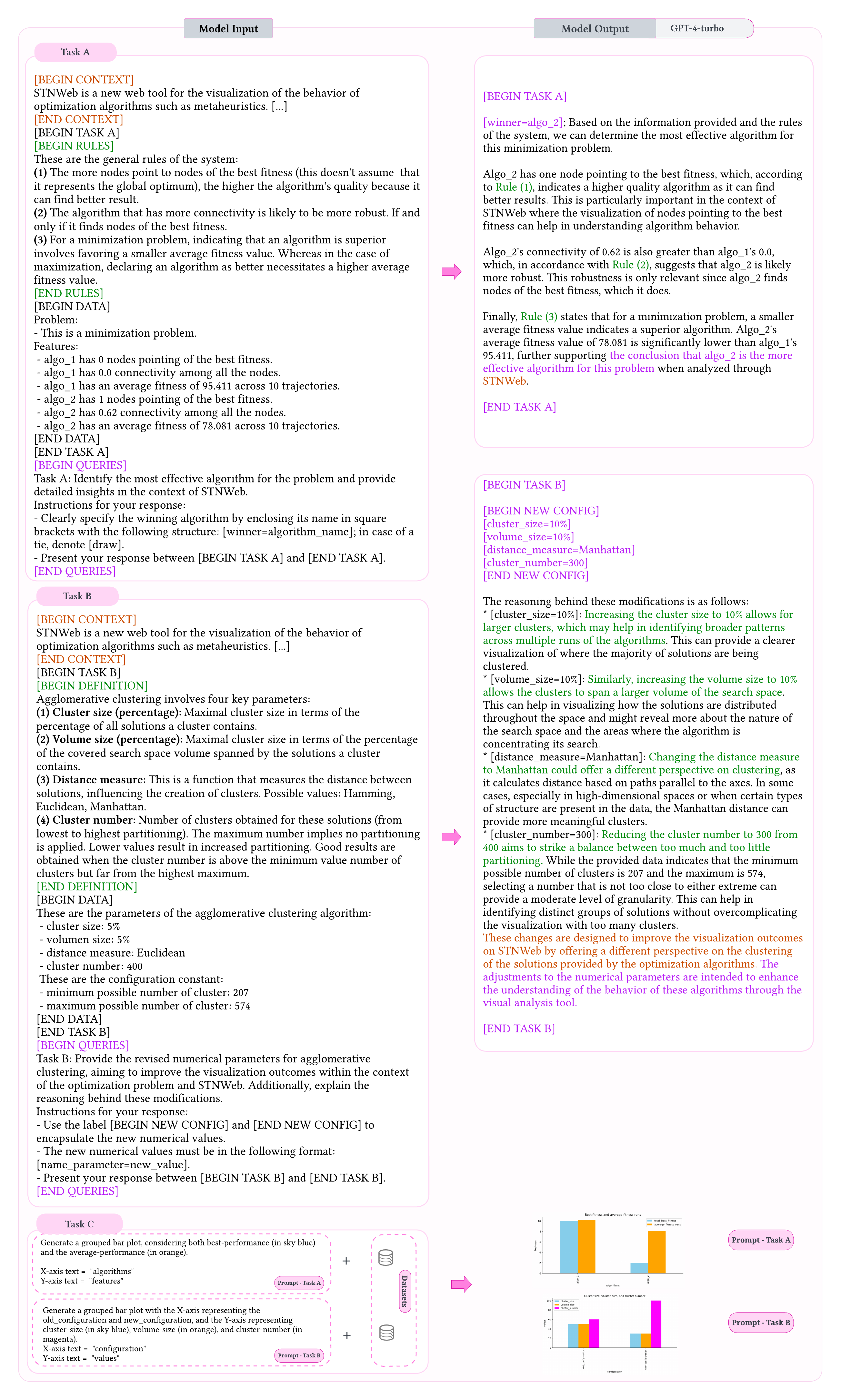}
  \caption{Prompts and model output (used LLM: GPT-4-turbo) concerning the STN graphic shown in Figure~\ref{fig:stn-example}.}
  \label{fig:examples-tasks}
\end{center}
\end{figure}

\subsection{Results}

This section provides the results in a summarized (quantitative) way. For a comprehensive presentation of every input (prompt) and output generated by each LLM, as well as to observe the variations in their responses, see the supplementary material, also available at \badge[purple]{\faicon{github} \href{https://github.com/camilochs/explainability-LLM-stnweb}{GitHub}}. Note that an example of the LLM responses is also provided in Figure~\ref{fig:examples-tasks}.

Table~\ref{table:results} provides the evaluation results for each task. Clearly, GPT-4-turbo excels for tasks~A and~B, handling both easy and hard prompts with greater proficiency. For the hard prompt of task A, demanding an evaluation of three algorithms, two of which with similar performance, a draw was the expected conclusion. However, GPT-4-turbo was the sole model to identify this accurately, in contrast to the consistent inaccuracies of other models. On the other hand, for the prompts classified as easy in task~A, the performance of all models was similar. Yet GPT-4-turbo stood out by delivering a better response, as per the human evaluators' judgment. Notably, for the easy prompt, GPT-4-turbo went straight to the correct answer (\textsc{[winner=algo\_2]}) and then began to justify this statement, diverging from the other models, which preferred to start with explanatory justifications before mentioning the winner. This indicates that GPT-4-turbo is more succinct and targeted in its responses. The partial score of $0.6$, instead of a perfect 1 for GPT-4-turbo in the easy prompt's human evaluation, arises from its correct identification of the winner (\textsc{[winner=algo\_2]}), coupled with its referencing of other algorithms in parentheses. This presentation is not stipulated by the \textsc{[QUERIES]} tag guidelines, which dictate that parenthesis should solely enclose the winner declaration: \textsc{[winner=algorithm\_name]} (refer to the supplementary material for details). In task~B, only GPT-4-turbo accurately followed the prescribed format for reporting on better configuration parameters for agglomerative clustering. The \textsc{[QUERIES]} tag specifies the format with the following instruction:

\begin{tcolorbox}[colback=purple!5, boxrule=0.5pt]
\textit{The new numerical values must be presented in the format: [name\_parameter=new\_value].}
\end{tcolorbox}

Other models did not comply with this format by replacing the “=” with “:” and omitting the parentheses. This can again be verified in the supplementary material. An example of a complete response from GPT-4-turbo is shown in Figure~\ref{fig:examples-tasks}, concerning the analysis of the STN graphic of Figure~\ref{fig:stn-example}.

For task~C, utilizing GPTs and Code Llama models via Chat2VIS does not yield significant distinctions. This could be due to the simplicity of our prompt and tabular data, which might not be complex enough to reveal any differences.

\begin{table}[!t]
\caption{LLMs evaluations for tasks.}\label{table:results}
\centering
\scriptsize
\resizebox{0.8\hsize}{!}{
\begin{tabular}{cc|lclc}
\toprule
\rowcolor[HTML]{ffd6f5}&  &\multicolumn{3}{c}{Evaluation} & \\\cmidrule{3-6}
\rowcolor[HTML]{ffd6f5}&  &\multicolumn{2}{c}{System } &\multicolumn{2}{c}{Human} \\\cmidrule{1-6}
\rowcolor[HTML]{ffd6f5}Task &\makecell{Prompt\\Type}   &LLM &Score ($\uparrow$) &LLM &Score ($\uparrow$) \\ \cmidrule{1-6}
\multirow{3}{*}{A} &\multirow{3}{*}{1/Easy} &GPT-4-turbo & \textbf{1} &GPT-4-turbo & \textbf{0.6}\\
 &  &mixtral-8x7b-instruct-v0.1 & 0.8 &mixtral-8x7b-instruct-v0.1 & 0.4 \\
 &  &tulu-2-dpo-70b & \textbf{1} &tulu-2-dpo-70b & 0 \\
\cmidrule{1-6}
\multirow{3}{*}{A} & \multirow{3}{*}{2/Hard} &GPT-4-turbo & \textbf{0.8} &GPT-4-turbo & \textbf{1}\\
 &  &mixtral-8x7b-instruct-v0.1 & 0.4 &mixtral-8x7b-instruct-v0.1 & 0 \\
 &  &tulu-2-dpo-70b & 0.1 &tulu-2-dpo-70b & 0\\
 \cmidrule{1-6}
\multirow{3}{*}{B} &\multirow{3}{*}{1/Easy} &GPT-4-turbo & \textbf{1} &GPT-4-turbo & \textbf{1} \\
 & &mixtral-8x7b-instruct-v0.1 & 0 &mixtral-8x7b-instruct-v0.1 & 0 \\
 & &tulu-2-dpo-70b & 0 &tulu-2-dpo-70b & 0 \\
 \cmidrule{1-6}
\multirow{3}{*}{B} &\multirow{3}{*}{2/Hard} &GPT-4-turbo & \textbf{0.6} &GPT-4-turbo & \textbf{1} \\
 & &mixtral-8x7b-instruct-v0.1 & 0 &mixtral-8x7b-instruct-v0.1 &  0 \\
 & &tulu-2-dpo-70b & 0 &tulu-2-dpo-70b &  0 \\
\bottomrule
\end{tabular}}

  \bigskip
  
\centering
\resizebox{0.5\hsize}{!}{
\begin{tabular}{clc}
\toprule
\rowcolor[HTML]{ffd6f5}  &\multicolumn{2}{c}{Human evaluation}  \\\cmidrule{1-3}
\rowcolor[HTML]{ffd6f5}Task  &LLM &Score ($\uparrow$) \\ \cmidrule{1-3}

\multirow{2}{*}{C}  &GPT-4  & \textbf{1} \\
  &CodeLlama-34b-Instruct-hf  & \textbf{1}\\
 
 \bottomrule
\end{tabular}
}
\end{table}

\section{Discussion}

When discussing the application of LLMs within tools from (and for) the optimization community, such as STNWeb, we consider three crucial aspects.

\begin{enumerate}
   \item LLMs' inherent \textbf{limitations and trustworthiness issues} demand users' awareness and close attention. Their inherent struggle with complex mathematical reasoning can hinder their ability to fully grasp intricate algorithmic concepts~\cite{lewkowycz2022solving, wei2023chainofthought}. Furthermore, the output quality of an LLM can be significantly impacted by the chosen model, potentially leading to inaccuracies or hallucinations~\cite{huang2023survey}. These suboptimal outcomes become more likely with prompts characterized by subjectivity and lack of structure. To address these challenges, effective guidance strategies and robust benchmarking methods are crucial. The recent emergence of various LLM prompting benchmark frameworks, such as those proposed in~\cite{zhu2023promptbench, lin2023llmeval, sun2023querydependent}, equips us with the tools to evaluate performance quality.
    \item \textbf{Open-source LLMs} are a crucial aspect of this discussion. Although proprietary models like GPT-4 outperform publicly available ones~\cite{brown2020language, Reynolds2021-eh}, as our results demonstrate, their performance can be enhanced through few-shot learning, refined prompt tuning strategies, and advanced prompt engineering methods~\cite{lester-etal-2021-power}. It is essential to focus future research on enhancing open-source models' performance and increasing their accessibility.
     \item \textbf{Explainability} is a vital factor in bridging the knowledge gap between experts and beginners. The use of prompt engineering in LLMs can facilitate tool usage and enhance explainability through multimodal LLMs that generate audio and images~\cite{yin2023survey}. However, our research indicates that the effectiveness of prompt engineering is highly dependent on the LLM's reasoning abilities. Specifically, our results were achieved using GPT-4, which exhibits a higher capacity for reasoning with simple rules. To enhance an LLM's explainability of a given STN graphic, a useful approach is to include, as part of the prompt, related articles previously published on the topic (e.g.~\cite{CHACONSARTORI2023100558, OCHOA2021107492}). This technique, known as Retrieval-Augmented Generation (RAG), allows access to external information (e.g.~via PDF files) that is not explicitly present in the prompts provided to the LLMs. In this way, the LLM may be able to generate more accurate responses~\cite{gao2024retrievalaugmented}.
\end{enumerate}

\section{Conclusion}

In this research, we leverage Large Language Models (LLMs) as assistants to automate the generation of text and plots, thereby reducing the prerequisite of extensive prior knowledge for utilizing STNWeb, a tool that facilitates the comparison of multiple optimization algorithms when applied to the same problem instance. Moreover, note that STNWeb allows for the analysis of algorithm behavior across discrete and continuous problem spaces. Our research underscores that with meticulous prompt engineering, LLMs can deliver reports with enhanced trustworthiness and explainability, despite their known struggles with complex mathematical reasoning.

Looking ahead, we aim to augment the set of algorithmic features extracted for prompt generation, thereby augmenting the explainability and utility of LLM-generated insights. Moreover, taking advantage of LLMs' impressive image analysis capabilities, we aim to create techniques that enable Multimodal LLMs to directly interpret graphs generated by STNWeb, thereby enriching the textual descriptions obtained on the basis of the prompts.


\subsection*{Acknowledgements}
C.~Chacón and C.~Blum were supported by grants TED2021-129319B-I00 and PID2022-136787NB-I00 funded by MCIN/AEI/10.13039/ 501100011033.

\printbibliography

\end{document}